\documentclass[letterpaper]{article} % DO NOT CHANGE THIS
\usepackage[preprint]{aaai2027}  % preprint: omit AAAI copyright notice (arXiv extended version)
\usepackage[hyphens]{url}  % DO NOT CHANGE THIS
\usepackage{graphicx} % DO NOT CHANGE THIS
\usepackage{natbib}  % DO NOT CHANGE THIS AND DO NOT ADD ANY OPTIONS TO IT
\usepackage{caption} % DO NOT CHANGE THIS AND DO NOT ADD ANY OPTIONS TO IT
\usepackage{booktabs}

\usepackage{tikz}
\usepackage{enumitem}
\usetikzlibrary{positioning, arrows.meta, fit, backgrounds, calc}

\providecommand{\Description}[1]{}

\title{Deploying a Control-Logic LLM Assistant in Air-Gapped Power Plants: Lessons in Source Authority and Bounded Retrieval}

\author{
    Yiwei Fu,
    Nicholas Gabriel
}
\affiliations{
    GE Vernova Advanced Research Center\\
    Niskayuna, NY, USA\\
    \{yiwei.fu, nicholas.gabriel\}@gevernova.com
}

\begin{document}

\maketitle

%% ============================================================
%% ABSTRACT
%% ============================================================
\begin{abstract}
Deploying Large Language Model (LLM) assistants in industrial operational technology environments poses constraints largely absent from cloud applications: site policies restrict engineering and operational data from leaving the plant, on-premises compute is limited, and incorrect causal attribution can mislead incident response.
We report lessons from deploying a control-logic evidence layer that underpins a broader on-premises operator assistance system at two gas power plants.
Two prototypes revealed where generative models added avoidable uncertainty and cost.
The first used a Vision Language Model (VLM) to reconstruct control topology from PDF printouts routinely exchanged by engineers.
Across the full 6{,}979-page document, extraction reached 0.51 connection precision and required 3.1 hours on four H200 GPUs; consultation with controls engineers then revealed a native structured export from the authoring system.
The second used an LLM tool-calling agent to plan repeatedly over eight predefined graph operations.
In a controlled benchmark of 40 template-generated questions, replacing this agent with explicit execution plans reduced mean token use by 20$\times$ and median latency by 3$\times$, with identical retrieved node sets across repeated runs.
The deployed design ingests the authoritative export into a versioned knowledge graph, executes bounded graph operations in fixed software, and confines the local LLM to fallback intent classification and evidence-grounded explanation.
At one site, the graph contains 221K nodes across seven controllers.
Our experience suggests a practical division of labor for high-consequence AI: LLMs interpret natural-language requests and explain results, while deterministic software constructs and retrieves technical evidence grounded in authoritative sources. Abstention, entity disambiguation, graph versioning, and read-only integration provide additional safeguards.
\end{abstract}

\begin{links}
  \par\noindent This is the extended version of a paper accepted to IAAI-27, including Appendices A--C.
\end{links}

%% ============================================================
%% 1. INTRODUCTION
%% ============================================================
\section{Introduction}\label{sec:intro}

Power plants are governed by control logic comprising tens of thousands of sensors, function blocks, and signal dependencies.
When a trip (emergency shutdown) or alarm occurs, an engineer traces backward through block diagrams and combines the resulting causal chain with event-time sensor data to determine what happened.
The tracing step alone can require hours of navigation through engineering tools and the expertise of a senior engineer.
In this context, we use \emph{causal tracing} to mean following cause-and-effect relationships explicitly encoded in executable control logic; attributing a particular event to one of those possible conditions additionally requires runtime measurements.

Deploying a Large Language Model (LLM)-based AI assistant for this task requires understanding of the exact control logic of a plant, not just general control knowledge, because no two plants have identical configurations.
Plant operational technology (OT) networks commonly prohibit sending live data to the public internet, ruling out cloud-hosted models.
The assistant must run on constrained on-premises hardware, preserve exact signal paths, fail safely on unsupported or ambiguous requests, and remain synchronized with revised control configurations.
Together, these constraints call for a local, versioned representation that preserves exact controller dependencies and supports deterministic, bounded traversal. We therefore transform the plant's control logic into a knowledge graph (KG) layer for on-premises LLM-assisted causal tracing.

These requirements make component allocation a deployment decision.
During development, we first used a Vision Language Model (VLM) to reconstruct topology from the routinely exchanged PDF printout.
After it failed at full-document scale, we consulted controls engineers about the underlying source and learned that the authoring tool provided a native structured export.
A second prototype used an LLM tool-calling agent to plan repeatedly over a closed set of graph operations.
We therefore ingest available structure from its authoritative source, implement enumerable operations as bounded code, and limit LLMs to interpreting natural-language requests and explaining retrieved evidence.
This task-boundary rule does not reject either technology: VLMs remain useful when no structured source exists, and agents when decomposition is genuinely open-ended.

Prior work does not address this combination directly: industrial KGs rarely expose executable block-level logic, while agentic graph systems target broader query spaces (Section~\ref{sec:related}).

We report how these constraints shaped a system now deployed at two gas power plants after validation on internal test systems.
Its construction pipeline ingests the control system's native \emph{Block Pin Report} and builds a property graph containing execution hierarchy, signal flow, and block equations.
Its query pipeline maps natural-language requests to a bounded set of predefined operations, resolves entities, executes native graph algorithms, and asks a local LLM to explain the retrieved evidence.
Here, \emph{deterministic} describes graph construction and execution after a query type and its entities have been selected; fallback intent classification and answer wording remain model-generated.
Event-specific diagnosis remains grounded in runtime measurements and engineering judgment.
Our contributions to deployment practice are:
\begin{enumerate}[nosep]
  \item A deployment-derived rule for assigning tasks to generative models versus deterministic software, supported by quantified accounts of an abandoned vision-based extraction pipeline and tool-calling retrieval agent.
  \item A deployed control-logic evidence pipeline that builds a versioned graph from an authoritative engineering export and maps natural-language requests to bounded execution plans, together with a controlled comparison against the agent it replaced.
  \item Operational practices for air-gapped industrial AI: local inference, explicit resource bounds, out-of-scope abstention, entity disambiguation, read-only integration, and a clear boundary between dependency tracing and event-specific diagnosis from runtime data.
\end{enumerate}

%% ============================================================
%% 2. RELATED WORK
%% ============================================================
\section{Related Work}\label{sec:related}

Manufacturing knowledge graphs usually fuse lifecycle, equipment, process, and event data~\cite{buchgeher2021knowledge,ringsquandl2017knowledge,ringsquandl2017event} or derive structure from standards such as AutomationML~\cite{grangel2018seamless}.
Diagnostic KGs have also matched patterns in historical service alerts~\cite{wang2024kgroot}.
These representations do not expose the executable block-level dependencies needed to trace a protection signal through controller logic.
The closest work extracts alarm-related logic from PLC code before statistical root-cause learning~\cite{wen2024alarm}; we instead represent complete production-scale control logic as a directly traversable evidence source.

Natural-language access to structured data commonly uses text-to-SQL or text-to-Cypher~\cite{liu2025survey,ozsoy2025text2cypher}, while GraphRAG and Think-on-Graph delegate parts of multi-hop retrieval to model reasoning~\cite{edge2024local,sun2023think}.
Multi-Agent GraphRAG adds schema validation, entity verification, and query refinement, yet reports 77\% accuracy in its best benchmark configuration~\cite{gusarov2025multi}.
These approaches target broader, open-ended query spaces.
Our deployed interface instead exposes eight known operations, making traversal expressible as fixed plans; generative planning adds a variable failure mode without useful expressivity for this interface.
Section~\ref{sec:exp-pipeline} evaluates that task-specific choice rather than claiming that bounded execution replaces agents for open-ended work.

Root-cause methods infer graphs from monitoring metrics, anomaly propagation, or industrial time series~\cite{chen2014causeinfer,han2025root,wen2022control}; LLM agents have also shown limited performance when diagnosing incidents from raw observations~\cite{roy2024exploring,xu2025openrca}.
These methods are complementary: runtime data establishes what occurred during an event, whereas our graph exposes the programmed dependencies through which an observed condition can propagate.

%% ============================================================
%% 3. DEPLOYED EVIDENCE PIPELINE
%% ============================================================
\section{Deployed Evidence Pipeline}\label{sec:construction}

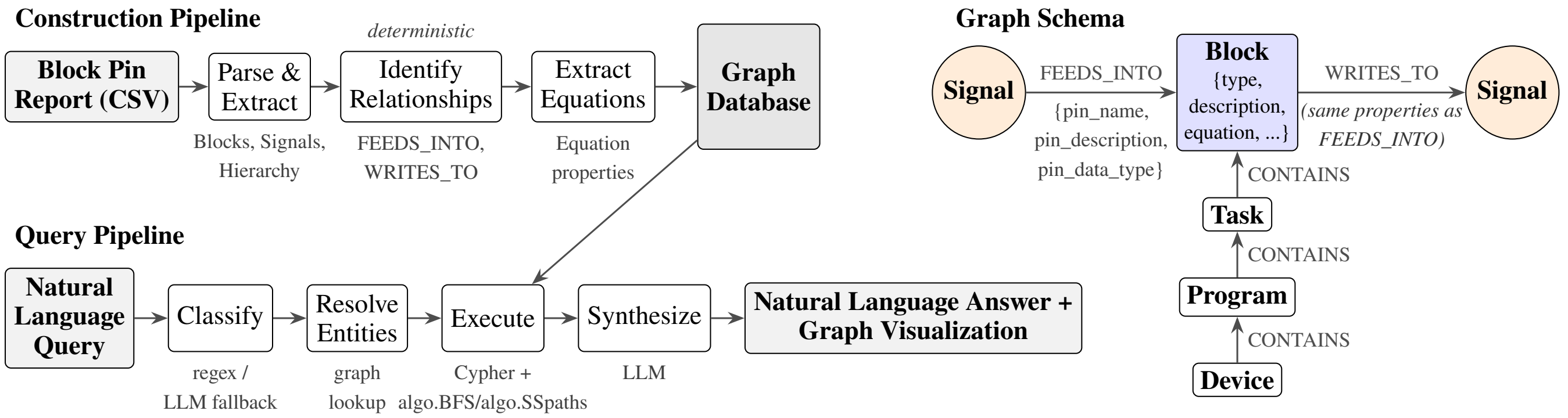
\begin{figure*}[th]
  \centering
  \resizebox{\textwidth}{!}{%
  \begin{tikzpicture}[
    >=Stealth,
    box/.style={
      draw, rounded corners=2pt, minimum height=0.8cm,
      font=\footnotesize, align=center, inner sep=3pt,
    },
    iobox/.style={box, fill=gray!10, font=\footnotesize\bfseries},
    stepbox/.style={box, fill=white},
    dbbox/.style={box, fill=gray!20, font=\footnotesize\bfseries,
                  minimum height=1.5cm, minimum width=1.3cm},
    sub/.style={font=\scriptsize, text=gray!50!black, align=center},
    tag/.style={font=\scriptsize\itshape, text=gray!50!black},
    arr/.style={->, semithick, gray!60!black},
    plabel/.style={font=\footnotesize\bfseries},
    schnode/.style={draw, inner sep=2.5pt, font=\footnotesize\bfseries},
    schlabel/.style={font=\scriptsize, text=black!70},
  ]

  % === Construction Pipeline (top) ===
  \node[plabel, anchor=west] at (0,0) (clabel) {Construction Pipeline};

  \node[iobox, below=0.1cm of clabel.south west, anchor=north west]
    (csv) {Block Pin\\Report (CSV)};
  \node[stepbox, right=0.35cm of csv] (parse) {Parse \&\\Extract};
  \node[stepbox, right=0.35cm of parse] (rels) {Identify\\Relationships};
  \node[stepbox, right=0.35cm of rels] (eqns) {Extract\\Equations};
  \node[dbbox, right=0.5cm of eqns] (db) {Graph\\Database};

  \draw[arr] (csv) -- (parse);
  \draw[arr] (parse) -- (rels);
  \draw[arr] (rels) -- (eqns);
  \draw[arr] (eqns) -- (db);

  \node[sub, below=1pt of parse] {Blocks, Signals,\\Hierarchy};
  \node[sub, below=1pt of rels] {FEEDS\_INTO,\\WRITES\_TO};
  \node[sub, below=1pt of eqns] {Equation\\properties};
  \node[tag, above=1pt of rels] {deterministic};

  % === Query Pipeline (bottom, left-aligned with construction) ===
  \node[plabel, anchor=west] at (0,-2.6) (qlabel) {Query Pipeline};

  \node[iobox, below=0.1cm of qlabel.south west, anchor=north west]
    (query) {Natural\\Language\\Query};
  \node[stepbox, right=0.4cm of query] (classify) {Classify};
  \node[stepbox, right=0.4cm of classify] (resolve) {Resolve\\Entities};
  \node[stepbox, right=0.4cm of resolve] (execute) {Execute};
  \node[stepbox, right=0.4cm of execute] (synth) {Synthesize};
  \node[iobox, right=0.4cm of synth] (answer) {Natural Language Answer +\\Graph Visualization};

  \draw[arr] (query) -- (classify);
  \draw[arr] (classify) -- (resolve);
  \draw[arr] (resolve) -- (execute);
  \draw[arr] (execute) -- (synth);
  \draw[arr] (synth) -- (answer);

  % DB -> Execute
  \draw[arr] (db) -- (execute);

  \node[sub, below=1pt of classify] {regex /\\LLM fallback};
  \node[sub, below=1pt of resolve] {graph\\lookup};
  \node[sub, below=1pt of execute] {Cypher +\\algo.BFS/algo.SSpaths};
  \node[sub, below=1pt of synth] {LLM};

  % === Graph Schema (right 1/3, top-aligned with Construction Pipeline) ===
  \node[plabel, anchor=west] at ([xshift=1.5cm]db.east |- clabel.west) (stitle) {Graph Schema};

  % Signal -> Block -> Signal (horizontal)
  \node[schnode, circle, fill=orange!15,
        below right=0.2cm and 0.0cm of stitle.south west] (s1) {Signal};
  \node[schnode, rounded corners=2pt, fill=blue!12, right=1.8cm of s1, align=center]
    (blk) {Block\\[-1pt]{\scriptsize\mdseries \{type,}\\[-1pt]{\scriptsize\mdseries description,}\\[-1pt]{\scriptsize\mdseries equation, ...\}}};
  \node[schnode, circle, fill=orange!15, right=2cm of blk] (s2) {Signal};

  % FEEDS_INTO: name above, properties below
  \draw[->, semithick, black!70] (s1) -- (blk)
    node[midway, above, schlabel] {FEEDS\_INTO}
    node[midway, below, sub] {\{pin\_name,\\pin\_description,\\pin\_data\_type\}};
  % WRITES_TO: name above, note same properties
  \draw[->, semithick, black!70] (blk) -- (s2)
    node[midway, above, schlabel] {WRITES\_TO}
    node[midway, below, sub, text=gray!40!black, font=\scriptsize\itshape] {(same properties as\\FEEDS\_INTO)};

  % Hierarchy: vertical, directly south of Block
  \node[schnode, rounded corners=2pt, below=0.55cm of blk] (task) {Task};
  \node[schnode, rounded corners=2pt, below=0.55cm of task] (prog) {Program};
  \node[schnode, rounded corners=2pt, below=0.55cm of prog] (dev) {Device};

  \draw[->, semithick, black!70] (dev) -- (prog)
    node[midway, right, schlabel] {CONTAINS};
  \draw[->, semithick, black!70] (prog) -- (task)
    node[midway, right, schlabel] {CONTAINS};
  \draw[->, semithick, black!70] (task) -- (blk)
    node[midway, right, schlabel] {CONTAINS};

  \end{tikzpicture}
  }
  \caption{Deployed evidence architecture. \emph{Left:} the construction pipeline maps a \emph{Block Pin Report} to a knowledge graph; the query pipeline classifies a request, resolves entities, executes a bounded graph operation, and synthesizes an explanation. \emph{Right:} the schema combines a bipartite Signal--Block flow graph with a four-level execution hierarchy.}
  \Description{System architecture diagram showing two pipelines (graph construction and query processing) on the left, and the graph schema with Signal, Block, Device, Program, and Task nodes on the right.}
  \label{fig:architecture}
\end{figure*}

The deployed layer has two pipelines (Figure~\ref{fig:architecture}): a \emph{graph construction} pipeline that maps a structured control logic export to a knowledge graph, and a \emph{query} pipeline that processes natural-language diagnostic questions against that graph.
The graph acts as an inspectable contract between the authoritative source and the language interface: construction preserves source relationships, and query code returns an evidence payload that answer synthesis cannot modify.
The construction pipeline ingests a \emph{Block Pin Report} CSV, extracts entities and relationships in a single pass, and loads the resulting graph into an in-memory graph database.
The query pipeline (Section~\ref{sec:query}) classifies incoming questions, resolves entity references, executes Cypher queries, and synthesizes a natural-language answer, using at most two LLM calls per query.

%% ----------------------------------------------------------
%% 3.1 Industrial Control Logic (domain primer)
%% ----------------------------------------------------------
\subsection{Control Logic and Source Authority}\label{sec:background}

A gas turbine controller executes \emph{control logic}: a large network of \emph{function blocks} that runs cyclically, typically every few tens of milliseconds. Each block is a small typed operator---a Boolean gate, an arithmetic or equation block, a comparator, a timer, or a sequencer---with named input and output \emph{pins}. Blocks are wired together by \emph{signals}: a signal carries the value of one block's output pin to the input pins of other blocks, forming a directed signal-flow network. An \emph{equation block} stores a Boolean or arithmetic expression over its input pins; for example, \texttt{(A+D)*\textasciitilde B*C} is true when \texttt{A} or \texttt{D} holds, \texttt{B} is false, and \texttt{C} is true, where \texttt{+}, \texttt{*}, and \texttt{\textasciitilde} denote logical OR, AND, and NOT. Variables in these expressions are pin names local to the block, which is why the same expression on two different blocks can refer to entirely different signals.

Certain outputs are \emph{protection} signals: when a protection signal transitions to true, the controller initiates a \emph{trip}, an automatic protective shutdown. Backward tracing reveals the programmed conditions that can produce that signal; identifying which condition occurred in a particular event additionally requires runtime measurements. The GE Vernova Mark~VIe ToolboxST engineering tool exports the block-and-pin structure of a controller as a flat CSV \emph{Block Pin Report}, the structured source our construction pipeline ingests.

%% ----------------------------------------------------------
%% 3.2 Parsing and Entity Extraction
%% ----------------------------------------------------------
\subsection{Construction from the Native Export}\label{sec:parsing}

The input to the graph construction pipeline is the \emph{Block Pin Report}, a structured CSV export from the GE Vernova Mark~VIe ToolboxST control system engineering tool.
Each row represents a single pin on a single function block, with columns including \texttt{Block}, \texttt{Block Type}, \texttt{Usage} (Input, Output, or Const), \texttt{Connection} (signal name), \texttt{Task}, and \texttt{Program}, among others.
A Block Pin Report for a typical heavy-duty gas turbine controller contains more than 100{,}000 rows, one per pin across all programs, tasks, and blocks; exact counts vary across models and site-specific configurations.
Although our implementation uses the GE Vernova Mark~VIe export format, the represented function blocks, typed interfaces, logical connections, and execution contexts align with concepts in IEC~61131-3~\cite{iec61131-3:2013} and with common practice in distributed control system (DCS) and programmable logic controller (PLC) platforms.

Entity extraction proceeds in a single pass over the CSV.
\textbf{Blocks} are identified by a composite key \texttt{Device|Program|Task|Block} that disambiguates blocks with the same name in different execution contexts; in the evaluated unit, for example, \texttt{RUNG\_1} appears in more than 150 distinct program--task contexts.
After deduplication, the rows yield tens of thousands of unique block nodes, each storing its type, description, execution order, and locator path as properties.
\textbf{Signals} are extracted from non-empty \texttt{Connection} values on non-constant pins.
A signal represents a named data path that connects the output of one block to the inputs of others.
% Signal names are deduplicated across the entire device, producing a comparable number of unique signal nodes.

The CSV also encodes a four-level execution hierarchy: each device contains programs, each program contains tasks, and each task contains blocks.
We materialize this hierarchy as \textsc{Contains} edges derived from the \texttt{Program} and \texttt{Task} columns.
This hierarchy supports scope-constrained queries and provides context for entity resolution in the query pipeline (Section~\ref{sec:query}).

The full construction pipeline completes in under 10~seconds for the evaluated gas turbine controller and under 30~seconds for the seven-controller plant on a laptop, with most time spent on CSV I/O.
Section~\ref{sec:relationships_equations} describes how signal-flow edges and equations are derived from the remaining columns.

%% ----------------------------------------------------------
%% 3.3 Relationship and Equation Processing
%% ----------------------------------------------------------
\subsection{Preserving Signal Flow and Equations}\label{sec:relationships_equations}

Signal-flow edges are derived directly from the \texttt{Usage} and \texttt{Connection} columns.
A row with \texttt{Usage=Input} and a non-empty \texttt{Connection} produces a \textsc{Feeds\_Into} edge from the named signal to the block; a row with \texttt{Usage=Output} and a non-empty \texttt{Connection} produces a \textsc{Writes\_To} edge from the block to the named signal.
Each edge carries pin metadata (name, description, data type) as properties, preserving the interface semantics.
A single signal may feed multiple pins on the same block (e.g., a temperature signal connected to both a comparison input and a display input), producing multiple \textsc{Feeds\_Into} edges that preserve pin-level connectivity.
Edges are deduplicated on the (signal, block, pin) triple to remove duplicate rows without collapsing distinct pin connections.

Equation extraction targets blocks with a constant pin named \texttt{EQN}, whose \texttt{Connection} field contains the equation string (e.g., \texttt{I1+I2*I3} or \texttt{(A*B*C)+(\textasciitilde E*D)}).
A regular expression extracts symbolic variable names, which correspond to pins on the same block and identify the inputs that participate in the expression.
In our schema, equations are stored as properties on Block nodes and interpreted by the LLM during answer synthesis.
An alternative schema we explored materialized equations as explicit \textsc{Logically\_Feeds\_Into} edges between Pin nodes, enabling pure graph-based intra-block traversal without LLM involvement; Section~\ref{sec:design-construction} discusses this trade-off.

%% ----------------------------------------------------------
%% 3.4 Design Decisions and Trade-offs
%% ----------------------------------------------------------
\subsection{Schema Decisions for Bounded Traversal}\label{sec:design-construction}

\paragraph{Why model pins as edge properties rather than nodes}
An alternative schema models pins as first-class nodes, producing a 4-hop path for each signal-to-signal step (Signal$\to$Pin$\to$Block$\to$Pin$\to$Signal).
Since native graph algorithms operate on hop counts, this doubled effective distances and complicated depth limits.
By folding pin metadata into edge properties, the deployed schema reduces each step to 2~hops (Signal$\to$Block$\to$Signal), forming a clean bipartite structure that BFS and shortest-path algorithms consume directly.
Equations are similarly stored as block properties rather than materialized as inter-pin edges. Topological path discovery therefore treats connected inputs as dependencies, while the equation is retained for explanation and event-time interpretation.

\paragraph{Why a graph database}
Control logic has two orthogonal relationship axes: a vertical execution hierarchy (Device $\to$ Program $\to$ Task $\to$ Block) and a lateral signal-flow graph (Signal $\leftrightarrow$ Block).
A backward trace may cross very different numbers of blocks, so the implementation needs variable-depth BFS and shortest-path operations over both axes.
Recursive relational queries could express these operations, but a property graph directly represents both relationship types while retaining typed pin metadata on edges.
We use FalkorDB, an open-source in-memory graph database; the schema and Cypher patterns are portable, although backend-specific graph procedures require adaptation.

%% ============================================================
%% 4. BOUNDED QUERY EXECUTION
%% ============================================================
\section{Bounded Query Execution}\label{sec:query}

The query pipeline maps a natural-language request onto a bounded set of graph operations rather than asking an LLM to generate Cypher or plan a traversal.
Each request passes through classification, entity resolution, execution, and answer synthesis.
Classification may use an LLM fallback, and synthesis always uses an LLM; entity resolution and graph execution are deterministic.
Thus each request makes at most two model calls, while the retrieved nodes, edges, and paths are produced by fixed code under explicit resource bounds.

%% ----------------------------------------------------------
%% 4.1 Classification
%% ----------------------------------------------------------
\subsection{Classification}\label{sec:query-classification}

The classifier recognizes eight in-scope query types: block or signal inspection (explore\_block, explore\_signal, signal\_producers, signal\_consumers) and dependency tracing (trace\_upstream, trace\_downstream, trace\_between, subgraph).
A no-LLM fast path extracts composite block identifiers and signal names, then applies phrase rules for intents including ``what drives,'' ``feeds into,'' and ``path between.''
Otherwise, an LLM selects one of the eight types or an explicit unsupported-request outcome.
The system abstains on unsupported requests rather than forcing them onto the nearest graph operation.

%% ----------------------------------------------------------
%% 4.2 Entity Resolution
%% ----------------------------------------------------------
\subsection{Entity Resolution}\label{sec:query-resolution}

After classification, the pipeline resolves all extracted names in parallel using exact \texttt{Signal.name}, exact \texttt{Block.name}, and composite \texttt{Block.block\_id} matches, with containment matching for composite identifiers.
A missing entity stops execution with a not-found response.
When a short block name maps to multiple execution contexts, the system returns the candidate scopes and asks the user to disambiguate instead of choosing one silently.
Centralizing this step avoids the repeated and potentially inconsistent lookups observed when a tool-calling agent decides for itself whether each name denotes a signal or block.

%% ----------------------------------------------------------
%% 4.3 Execution
%% ----------------------------------------------------------
\subsection{Execution}\label{sec:query-execution}

Each query type has a dedicated Cypher executor that returns both an evidence payload and visualization nodes and edges.
Exploration operations retrieve adjacent signals and block metadata.
Trace operations use native BFS or shortest-path procedures and enrich discovered blocks with their type, equation, pins, and connected signals in the same retrieval operation.
The implementation caps trace depth ($\le$10~hops), BFS frontier size ($\le$8), and total node count ($\le$150), and bounds shortest-path length ($\le$16).
The evidence payload records whether any bound truncated the result.
Entity resolution and retrieval therefore require two database round trips for the evaluated query types.

%% ----------------------------------------------------------
%% 4.4 Synthesis
%% ----------------------------------------------------------
\subsection{Evidence-Grounded Answer Generation}\label{sec:query-synthesis}

The final stage serializes the retrieved evidence into query-specific context: block questions include equations and pin-to-signal mappings, trace questions include ordered dependency layers, and path questions include the node sequence.
The LLM explains this evidence in natural language but cannot alter the graph payload shown to the user.
Its prose remains LLM-generated and may vary or contain errors; the deterministic guarantee applies to the graph payload after intent and entity selection, not to the wording of the answer. Confining synthesis to retrieved evidence narrows the opportunity for unsupported claims.
Removing traversal planning from the model reduces context and computation enough for the evaluated local model to run on a single on-premises server (Section~\ref{sec:deployment}).

%% ============================================================
%% 5. TWO MODEL-FIRST PROTOTYPES
%% ============================================================
\section{Two Model-First Prototypes}\label{sec:experiments}

The deployed pipeline replaced two earlier model-first prototypes.
Their failure modes differed, but the architectural cause was the same: the VLM inferred structure we later learned the authoring system could export, while the agent inferred an operation sequence that the application could enumerate.
The following accounts explain why we replaced them and extract decision rules; they do not argue against VLMs or agents for open-ended tasks.

We first establish the scale at which these design decisions operate.
We evaluate the system on control logic from a gas power plant equipped with seven GE Vernova controllers.
Experiments use one of the gas turbine controllers as the primary unit, whose Block Pin Report contains 168{,}220~rows and produces the graph summarized in Table~\ref{tab:graph-stats}.
To assess plant-level scale, we also construct the full graph spanning all seven controllers (gas turbine, safety controller, steam turbine, heat recovery steam generator, and balance-of-plant units); the number of controllers varies across sites.
Construction time for the full plant graph remains under 30~seconds on a laptop.

\begin{table}[t]
  \caption{Knowledge graph statistics. Single GT Controller is the primary evaluation unit; Full Plant covers seven controllers at one site. Total nodes also include Device, Program, and Task hierarchy nodes.}
  \label{tab:graph-stats}
  \centering
  \small
  \begin{tabular}{lrr}
    \toprule
    Metric & Single GT & Full Plant \\
    \midrule
    Devices             & 1       & 7 \\
    Block nodes         & 28{,}189 & 97{,}005 \\
    Signal nodes        & 29{,}721 & 120{,}742 \\
    Total nodes         & 58{,}644 & 221{,}406 \\
    Total edges         & 116{,}107 & 402{,}240 \\
    \midrule
    CSV rows            & 168{,}220 & 812{,}547 \\
    Construction time   & $<10$ s     & $<30$ s \\
    \bottomrule
  \end{tabular}
\end{table}

\subsection{Lesson 1: Find the Source Before Extracting}\label{sec:exp-vlm}

Our first graph construction prototype began from the controller's 6{,}979-page control logic PDF, the artifact routinely exchanged by engineers. We exported each page as an image, sent it to a VLM, and parsed the extracted entities and connections into a graph. The structured export was not part of the usual exchange workflow and was initially unknown to the project team; moreover, outputs from individual pages looked plausible enough to suggest that scaling the prototype might work.

Full-document processing exposed why that apparent success was misleading. The PDF was unsuitable as the source of record for three structural reasons: 65\% of its pages were not diagrams and yielded spurious entities; labels visible in the printout did not consistently match internal pin and signal identifiers; and page-local extraction could not recover connections that crossed page boundaries. Downstream stages in our pipeline consumed rather than independently revalidated the topology, so every trace and explanation inherited these errors. We therefore abandoned the VLM path and consulted controls engineers to identify an authoritative machine-readable source, which led us to the native \emph{Block Pin Report}.

Qwen~3.5-397B-A17B~\cite{qwen3.5} reached 0.51 relaxed connection precision and 36\% exact equation accuracy over the full document, compared with 75\% equation accuracy on the 100-page sample, and required 3.1 hours on four H200 GPUs (approximately \$124). Gemini~3.1 Pro~\cite{gemini} reached 0.24 connection precision and 4\% equation accuracy on the shared sample. By contrast, the native Block Pin Report directly specifies block identity, pin direction, and connections and was ingested in under 10 seconds for the same controller. Because export records cannot be completely aligned with individual pages, the VLM study measures precision but not recall; it characterizes the false-positive burden of the rejected prototype rather than constituting a controlled comparison of equivalent inputs. Appendix~\ref{app:vlm} gives the full protocol and results.

The transferable lesson is procedural rather than model-specific: source discovery should precede model selection. Before designing extraction from an engineering document, teams should identify its authoring system, ask domain engineers about native exports and APIs, and determine whether the document is a primary source or a rendered view. This rule also applies to piping and instrumentation diagrams~\cite{alimin2025talking}, process-flow diagrams, and electrical single-line diagrams. A VLM may still recover information absent from structured sources, but it should not reconstruct relationships that those sources already encode.

%% ----------------------------------------------------------
%% 5.2 Lesson 2: Bounded vs Agentic Retrieval
%% ----------------------------------------------------------
\subsection{Lesson 2: Do Not Re-plan Enumerable Operations}\label{sec:exp-pipeline}

We compare the deployed classify-resolve-execute design with an earlier tool-calling prototype. Both use Qwen~3.5-35B-A3B-FP8 on an NVIDIA GB10 with 128~GB of unified memory, the same graph, and the same evidence-grounded synthesis prompt. The agent receives block, signal, path-tracing, and arbitrary-Cypher tools and decides which to call and when to stop; the deployed design selects a fixed executor after classification. This controlled comparison varies who plans retrieval, not the underlying data or answer model.

The workload contains 40 template-generated questions spanning the eight supported operations, sampled from a 6{,}860-node neighborhood around one alarm and run three times. Expected answers were generated from query-specific graph subgraphs and scored with DeepEval's LLM-based GEval rubric~\cite{confident2024deepeval,liu2023g}. This design supports a controlled comparison but is not a broad natural-language benchmark or a substitute for human validation; Appendix~\ref{app:query-eval} gives the protocol and rubric.

The bounded pipeline achieved a mean LLM-judged answer-correctness score of 94\%, compared with 85\% for the agent. It reduced median latency from 14.2 to 4.8 seconds, p95 latency from 108.6 to 12.7 seconds, and mean token use from 24K to 1.2K per query (Table~\ref{tab:pipeline-comparison}). Its retrieved node sets were identical across repeated runs (Jaccard 1.0), whereas the agent averaged 0.61 because its tool sequence varied. The agent issued 0--31 graph calls and occasionally answered without retrieval; the bounded pipeline always retrieved evidence in two database round trips. On a single on-premises server, the agent's 108.6-second p95 and variable call count made response time and capacity depend on model stopping behavior. Fixed plans instead bound the number of model and database calls independently of trace depth.

The resulting decision rule is to enumerate before delegating: when valid operations, parameters, and resource bounds can be specified in advance, classify once and execute a fixed plan. Agentic planning remains appropriate when valid task decomposition cannot be enumerated before execution.

\begin{table}[t]
  \caption{Controlled comparison over 40 template-generated queries, three runs each. Answer correctness is LLM-judged; Jaccard measures cross-run retrieved-node overlap.}
  \label{tab:pipeline-comparison}
  \centering
  \small
  \begin{tabular}{lrr}
    \toprule
    Metric & Tool-calling & Bounded \\
    \midrule
    Answer correctness & 85\% & 94\% \\
    Latency, p50 / p95 (s) & 14.2 / 108.6 & 4.8 / 12.7 \\
    Tokens / query, mean & 24K & 1.2K \\
    DB round-trips, mean & 9.1 & 2.0 \\
    Retrieval overlap (Jaccard) & 0.61 & 1.0 \\
    \bottomrule
  \end{tabular}
\end{table}

%% ============================================================
%% 6. PRODUCTION DEPLOYMENT AND ASSURANCE
%% ============================================================
\section{Production Deployment and Assurance}\label{sec:deployment}

The control-logic query layer is in operational use for post-trip analysis and alarm investigation at two gas power plants following validation on internal test systems. Starting from a protection signal, it surfaces upstream blocks, signals, and activation conditions as an interactive dependency graph; runtime measurements are then needed to determine which possible condition occurred. Appendix~\ref{app:usecase} provides a worked example for \texttt{L86CBPS\_F\_A}. Deployment exposed requirements that are easy to miss in a laboratory prototype: offline operation, bounded resource use, safe handling of ambiguous requests, synchronization with changing control logic, and a clear boundary between decision support and plant control.

\subsection{Offline and Advisory by Design}

At the deployed sites, OT policies prevent data from leaving the plant, ruling out cloud-hosted models for this workload. The graph database, application, and local model therefore run on a single on-premises server. Moving traversal planning out of the LLM makes this modest deployment footprint practical and bounds model calls independently of path length. The system connects to the plant's DCS alarm server and historian in a read-only, advisory-only configuration: it presents evidence to engineers and operators but cannot write to the control system or take control action.

\subsection{Failure Containment and Configuration Lifecycle}

Natural-language interfaces create operational failure modes even when graph execution is deterministic. The deployed classifier can abstain on unsupported requests, and entity resolution asks for clarification when a short block name has multiple scoped matches. These controls prevent an apparently valid answer from being generated for the wrong operation or component. Control configurations also change over a plant's life. Each control logic graph is versioned and rebuilt from a new engineering export after configuration changes rather than mixing records from different revisions. Together with read-only integration, these practices keep the assistant advisory and make its evidence traceable to a specific source configuration.

\subsection{Retrospective Incident Replay}\label{sec:incident-replay}

\begin{figure}[t]
  \centering
  \begin{tikzpicture}[
    >=Stealth,
    box/.style={draw, rounded corners=2pt, minimum height=0.55cm,
                font=\ttfamily\bfseries\LARGE, align=center, inner sep=2.5pt},
    signode/.style={box, fill=blue!10},
    instrnode/.style={box, fill=blue!30},
    morenode/.style={box, rounded corners=5pt, fill=gray!12,
                     font=\normalsize\itshape},
    focusnode/.style={box, very thick, fill=red!60},
    arr/.style={->, semithick, black!65},
    scale=0.6, transform shape,
  ]

  % --- Upstream predecessors ---
  \node[signode]  (lk14p22)  at (-5.0, 0)  {LK14P22};
  \node[signode]  (lk14pr1)  at (-2.5, 0)  {LK14PR1};
  \node[morenode] (more)     at ( 0,   0)  {\, +23 more \,};
  \node[signode]  (tnk14ls2) at ( 2.5, 0)  {TNK14LS2};
  \node[signode]  (tnk14lx1) at ( 5.5, 0)  {TNK14LX1};

  % --- L14TV1 ---
  \node[instrnode] (l14tv1) at (0, -1.1) {L14TV1};

  \draw[arr] (lk14p22)  -- (l14tv1);
  \draw[arr] (lk14pr1)  -- (l14tv1);
  \draw[arr] (more)     -- (l14tv1);
  \draw[arr] (tnk14ls2) -- (l14tv1);
  \draw[arr] (tnk14lx1) -- (l14tv1);

  % --- Two sub-branches ---
  \node[signode] (l14hs_rsl) at (-1.6,  -2.2) {L14HS\_RSL};
  \node[signode] (l14ht_rsl) at ( 1.6,  -2.2) {L14HT\_RSL};

  \draw[arr] (l14tv1) -- (l14hs_rsl);
  \draw[arr] (l14tv1) -- (l14ht_rsl);

  % Left (HS) branch
  \node[instrnode] (move10) at (-1.6, -3.2) {MOVE\_10};
  \node[signode]   (l14hs)  at (-1.6, -4.2) {L14HS};
  \node[instrnode] (move3l) at (-1.6, -5.2) {MOVE\_3};

  \draw[arr] (l14hs_rsl) -- (move10);
  \draw[arr] (move10)    -- (l14hs);
  \draw[arr] (l14hs)     -- (move3l);

  % Right (HT) branch
  \node[instrnode] (move12) at (1.6, -3.2) {MOVE\_12};
  \node[signode]   (l14ht)  at (1.6, -4.2) {L14HT};
  \node[instrnode] (move8)  at (1.6, -5.2) {MOVE\_8};

  \draw[arr] (l14ht_rsl) -- (move12);
  \draw[arr] (move12)    -- (l14ht);
  \draw[arr] (l14ht)     -- (move8);

  % --- Branch outputs ---
  \node[signode] (pos_cv_dh) at (-4, -6.4)
    { 90CB1.POS\_CV\_DH};
  \node[signode] (l14hs_cbv) at ( 0,   -6.4) {L14HS\_CBV};
  \node[signode] (l14ht_cbv) at ( 3., -6.4) {L14HT\_CBV};

  %\draw[arr] (move3l) -- (pos_cv_dh);
  \draw[arr] (move3l) -- (l14hs_cbv);
  %\draw[arr] (move8)  -- (l14hs_cbv);
  \draw[arr] (move8)  -- (l14ht_cbv);

  % --- Main downstream chain (horizontal) ---
  \node[instrnode] (rung29)     at (0,  -7.5)  {RUNG\_29};

  \draw[arr] (l14ht_cbv)  -- (rung29);
  \draw[arr] (pos_cv_dh)  -- (rung29);
  \draw[arr] (l14hs_cbv)  -- (rung29);

  \node[signode]   (l86cb1ps_f) at (-5, -8.4) {L86CB1PS\_F};
  \node[instrnode] (move3m)     at (-1.8, -8.4) {MOVE\_3};
  \node[signode]   (l86cbps_f)  at ( 1.2, -8.4) {L86CBPS\_F};
  \node[instrnode] (move13)     at ( 4.3, -8.4) {MOVE\_13};

  \node[focusnode] (l86cbps_fa) at (0, -9.5) {L86CBPS\_F\_A};

  \draw[arr] (rung29)     -- (l86cb1ps_f);
  \draw[arr] (l86cb1ps_f) -- (move3m);
  \draw[arr] (move3m)     -- (l86cbps_f);
  \draw[arr] (l86cbps_f)  -- (move13);
  \draw[arr] (move13)     -- (l86cbps_fa);

  \end{tikzpicture}
  \caption{Subgraph returned by backward traversal from a trip alarm. Arrows retain the forward signal-flow direction from upstream inputs to \texttt{L86CBPS\_F\_A}.}
  \Description{Directed graph returned by a backward traversal from L86CBPS\_F\_A, with arrows showing forward signal flow from upstream inputs through control logic blocks to the alarm.}
  \label{fig:l86cbps}
\end{figure}
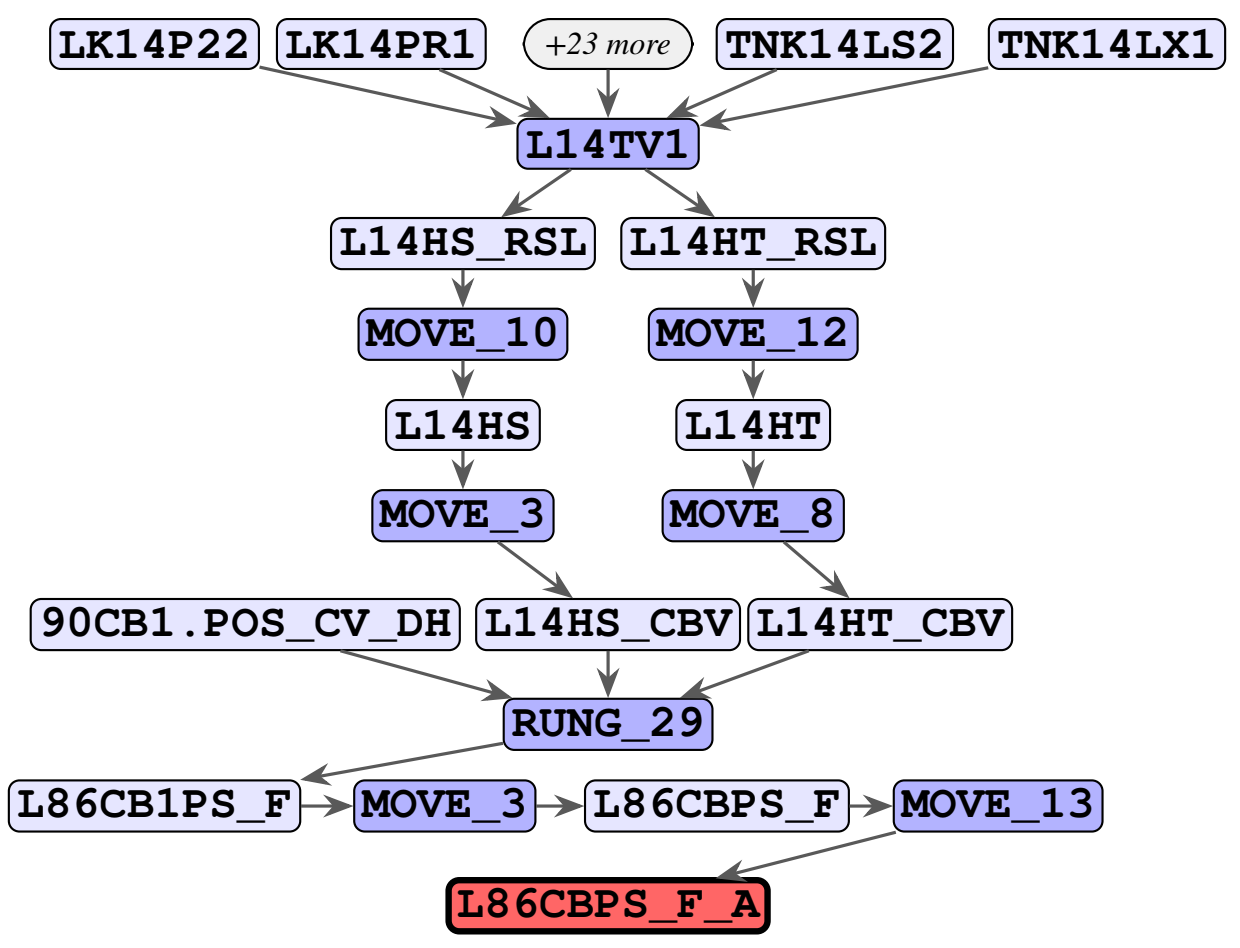

We retrospectively replayed an October~2025 compressor bleed valve (CBV) trip from a unit where the system is deployed. The protection signal \texttt{L86CBPS\_F\_A} denotes disagreement between commanded and measured valve position. A backward query surfaced its RUNG\_29 logic and the \texttt{90CB1.POS\_CV\_DH} position-deviation input (Figure~\ref{fig:l86cbps}; Appendix~\ref{app:usecase} gives the full query and response). A separate trip-analysis capability then evaluated archived historian values: the command fell from 100 to~40 while feedback remained near~97, confirming that the valve did not move before the trip. The replay demonstrates the intended division of labor---the graph assistant exposes the deterministic dependency structure, while runtime evidence establishes which condition occurred.

\subsection{Limitations}\label{sec:limitations}

Our evidence is subject to several limitations. Both evaluations use one vendor's control format, and the query benchmark uses a template-generated workload centered on one alarm. The LLM-generated answers are nondeterministic and were evaluated with an LLM judge rather than independent human review, although the underlying graph payload is preserved for inspection. Control logic and production code are proprietary and cannot be released. The reported field example is a single retrospective incident replay, and current site policies prevent the installations from exporting usage telemetry to a central service; we therefore report controlled technical measurements and deployment experience, not measured changes in operator response time or plant availability. Future work includes local privacy-preserving operational metrics and temporal reasoning over historian data.

\section{Conclusion}\label{sec:conclusion}

We deployed a control-logic evidence layer at two air-gapped gas power plants. In these cases, the ability of a VLM or LLM agent to perform a step did not make it the appropriate production mechanism: the first reconstructed topology from the familiar PDF before we identified the authoring system's native export, while the second re-planned eight enumerable graph operations. Replacing those steps with authoritative ingestion and bounded execution produced a 221{,}406-node plant graph and, in a controlled benchmark, reduced token use by 20$\times$ and median latency by 3$\times$ while returning identical node sets across runs. The resulting division of labor is practical for high-consequence industrial AI: LLMs interpret natural-language requests and explain results; deterministic software constructs and retrieves evidence grounded in authoritative sources; and runtime measurements plus engineering judgment establish what occurred during an event.

%% ============================================================
%% REFERENCES
%% ============================================================
\bibliography{references}

\begin{thebibliography}{23}
\providecommand{\natexlab}[1]{#1}

\bibitem[{Alimin et~al.(2025)Alimin, Goldstein, Balhorn, and
  Schweidtmann}]{alimin2025talking}
Alimin, A.~A.; Goldstein, D.~P.; Balhorn, L.~S.; and Schweidtmann, A.~M. 2025.
\newblock Talking like piping and instrumentation diagrams (p\&ids).
\newblock \emph{arXiv preprint arXiv:2502.18928}.

\bibitem[{Buchgeher et~al.(2021)Buchgeher, Gabauer, Martinez-Gil, and
  Ehrlinger}]{buchgeher2021knowledge}
Buchgeher, G.; Gabauer, D.; Martinez-Gil, J.; and Ehrlinger, L. 2021.
\newblock Knowledge graphs in manufacturing and production: a systematic
  literature review.
\newblock \emph{Ieee Access}, 9: 55537--55554.

\bibitem[{Chen et~al.(2014)Chen, Qi, Zheng, and Hou}]{chen2014causeinfer}
Chen, P.; Qi, Y.; Zheng, P.; and Hou, D. 2014.
\newblock Causeinfer: Automatic and distributed performance diagnosis with
  hierarchical causality graph in large distributed systems.
\newblock In \emph{IEEE INFOCOM 2014-IEEE Conference on Computer
  Communications}, 1887--1895. IEEE.

\bibitem[{Edge et~al.(2024)Edge, Trinh, Cheng, Bradley, Chao, Mody, Truitt,
  Metropolitansky, Ness, and Larson}]{edge2024local}
Edge, D.; Trinh, H.; Cheng, N.; Bradley, J.; Chao, A.; Mody, A.; Truitt, S.;
  Metropolitansky, D.; Ness, R.~O.; and Larson, J. 2024.
\newblock From local to global: A graph rag approach to query-focused
  summarization.
\newblock \emph{arXiv preprint arXiv:2404.16130}.

\bibitem[{Grangel-Gonz{\'a}lez et~al.(2018)Grangel-Gonz{\'a}lez, Halilaj,
  Vidal, Lohmann, Auer, and M{\"u}ller}]{grangel2018seamless}
Grangel-Gonz{\'a}lez, I.; Halilaj, L.; Vidal, M.-E.; Lohmann, S.; Auer, S.; and
  M{\"u}ller, A.~W. 2018.
\newblock Seamless integration of cyber-physical systems in knowledge graphs.
\newblock In \emph{Proceedings of the 33rd Annual ACM Symposium on Applied
  Computing}, 2000--2003.

\bibitem[{Gusarov et~al.(2025)Gusarov, Volkova, Khrulkov, Kuznetsov, Maslov,
  and Oseledets}]{gusarov2025multi}
Gusarov, A.; Volkova, A.; Khrulkov, V.; Kuznetsov, A.; Maslov, E.; and
  Oseledets, I. 2025.
\newblock Multi-Agent GraphRAG: A Text-to-Cypher Framework for Labeled Property
  Graphs.
\newblock \emph{arXiv preprint arXiv:2511.08274}.

\bibitem[{Han et~al.(2025)Han, Absar, Zhang, and Yuan}]{han2025root}
Han, X.; Absar, S.; Zhang, L.; and Yuan, S. 2025.
\newblock Root cause analysis of anomalies in multivariate time series through
  granger causal discovery.
\newblock In \emph{The Thirteenth International Conference on Learning
  Representations}.

\bibitem[{{International Electrotechnical Commission}(2013)}]{iec61131-3:2013}
{International Electrotechnical Commission}. 2013.
\newblock \emph{{IEC 61131-3:2013} Programmable controllers -- Part 3:
  Programming languages}.
\newblock International Electrotechnical Commission (IEC), Geneva, Switzerland.

\bibitem[{Ip and Vongthongsri(2026)}]{confident2024deepeval}
Ip, J.; and Vongthongsri, K. 2026.
\newblock {deepeval}.

\bibitem[{Kwon et~al.(2023)Kwon, Li, Zhuang, Sheng, Zheng, Yu, Gonzalez, Zhang,
  and Stoica}]{kwon2023efficient}
Kwon, W.; Li, Z.; Zhuang, S.; Sheng, Y.; Zheng, L.; Yu, C.~H.; Gonzalez, J.~E.;
  Zhang, H.; and Stoica, I. 2023.
\newblock Efficient Memory Management for Large Language Model Serving with
  PagedAttention.
\newblock In \emph{Proceedings of the ACM SIGOPS 29th Symposium on Operating
  Systems Principles}.

\bibitem[{Liu et~al.(2025)Liu, Shen, Li, Ma, Jiang, Zhang, Fan, Li, Tang, and
  Luo}]{liu2025survey}
Liu, X.; Shen, S.; Li, B.; Ma, P.; Jiang, R.; Zhang, Y.; Fan, J.; Li, G.; Tang,
  N.; and Luo, Y. 2025.
\newblock A Survey of Text-to-SQL in the Era of LLMs: Where are we, and where
  are we going?
\newblock \emph{IEEE Transactions on Knowledge and Data Engineering}.

\bibitem[{Liu et~al.(2023)Liu, Iter, Xu, Wang, Xu, and Zhu}]{liu2023g}
Liu, Y.; Iter, D.; Xu, Y.; Wang, S.; Xu, R.; and Zhu, C. 2023.
\newblock G-eval: NLG evaluation using gpt-4 with better human alignment.
\newblock In \emph{Proceedings of the 2023 conference on empirical methods in
  natural language processing}, 2511--2522.

\bibitem[{Ozsoy et~al.(2025)Ozsoy, Messallem, Besga, and
  Minneci}]{ozsoy2025text2cypher}
Ozsoy, M.~G.; Messallem, L.; Besga, J.; and Minneci, G. 2025.
\newblock Text2cypher: Bridging natural language and graph databases.
\newblock In \emph{Proceedings of the Workshop on Generative AI and Knowledge
  Graphs (GenAIK)}, 100--108.

\bibitem[{{Qwen Team}(2026)}]{qwen3.5}
{Qwen Team}. 2026.
\newblock {Qwen3.5}: Towards Native Multimodal Agents.

\bibitem[{Ringsquandl et~al.(2017{\natexlab{a}})Ringsquandl, Kharlamov,
  Stepanova, Lamparter, Lepratti, Horrocks, and
  Kr{\"o}ger}]{ringsquandl2017event}
Ringsquandl, M.; Kharlamov, E.; Stepanova, D.; Lamparter, S.; Lepratti, R.;
  Horrocks, I.; and Kr{\"o}ger, P. 2017{\natexlab{a}}.
\newblock On event-driven knowledge graph completion in digital factories.
\newblock In \emph{2017 IEEE International Conference on Big Data (Big Data)},
  1676--1681. IEEE.

\bibitem[{Ringsquandl et~al.(2017{\natexlab{b}})Ringsquandl, Lamparter,
  Lepratti, and Kr{\"o}ger}]{ringsquandl2017knowledge}
Ringsquandl, M.; Lamparter, S.; Lepratti, R.; and Kr{\"o}ger, P.
  2017{\natexlab{b}}.
\newblock Knowledge fusion of manufacturing operations data using
  representation learning.
\newblock In \emph{IFIP International Conference on Advances in Production
  Management Systems}, 302--310. Springer.

\bibitem[{Roy et~al.(2024)Roy, Zhang, Bhave, Bansal, Las-Casas, Fonseca, and
  Rajmohan}]{roy2024exploring}
Roy, D.; Zhang, X.; Bhave, R.; Bansal, C.; Las-Casas, P.; Fonseca, R.; and
  Rajmohan, S. 2024.
\newblock Exploring llm-based agents for root cause analysis.
\newblock In \emph{Companion proceedings of the 32nd ACM international
  conference on the foundations of software engineering}, 208--219.

\bibitem[{Sun et~al.(2023)Sun, Xu, Tang, Wang, Lin, Gong, Ni, Shum, and
  Guo}]{sun2023think}
Sun, J.; Xu, C.; Tang, L.; Wang, S.; Lin, C.; Gong, Y.; Ni, L.~M.; Shum, H.-Y.;
  and Guo, J. 2023.
\newblock Think-on-graph: Deep and responsible reasoning of large language
  model on knowledge graph.
\newblock \emph{arXiv preprint arXiv:2307.07697}.

\bibitem[{Team et~al.(2023)Team, Anil, Borgeaud, Alayrac, Yu, Soricut,
  Schalkwyk, Dai, Hauth, Millican et~al.}]{gemini}
Team, G.; Anil, R.; Borgeaud, S.; Alayrac, J.-B.; Yu, J.; Soricut, R.;
  Schalkwyk, J.; Dai, A.~M.; Hauth, A.; Millican, K.; et~al. 2023.
\newblock Gemini: a family of highly capable multimodal models.
\newblock \emph{arXiv preprint arXiv:2312.11805}.

\bibitem[{Wang, Qi, and Wu(2024)}]{wang2024kgroot}
Wang, T.; Qi, G.; and Wu, T. 2024.
\newblock KGroot: Enhancing root cause analysis through knowledge graphs and
  graph convolutional neural networks.
\newblock \emph{arXiv preprint arXiv:2402.13264}.

\bibitem[{Wen et~al.(2022)Wen, Yan, Liang, Wu, Zhou, and Yao}]{wen2022control}
Wen, C.-M.; Yan, Z.; Liang, Y.-C.; Wu, H.; Zhou, L.; and Yao, Y. 2022.
\newblock A control chart-based symbolic conditional transfer entropy method
  for root cause analysis of process disturbances.
\newblock \emph{Computers \& Chemical Engineering}, 164: 107902.

\bibitem[{Wen et~al.(2024)Wen, Kunze, Wilch, Fay, and
  Vogel-Heuser}]{wen2024alarm}
Wen, Z.; Kunze, F.~C.; Wilch, J.; Fay, A.; and Vogel-Heuser, B. 2024.
\newblock Alarm Root-Cause Analysis Using an Alarm Logic Directed Graph
  Extracted From Control Software.
\newblock In \emph{2024 IEEE 3rd Industrial Electronics Society Annual On-Line
  Conference (ONCON)}, 1--8. IEEE.

\bibitem[{Xu et~al.(2025)Xu, Jiang, Hua, Li, Wang, and Li}]{xu2025openrca}
Xu, Z.; Jiang, Z.; Hua, Y.; Li, J.; Wang, C.; and Li, Z. 2025.
\newblock {OpenRCA}: Can LLM Agents Outperform Humans in Root Cause Analysis of
  Cloud Incidents?
\newblock \emph{arXiv preprint arXiv:2502.15735}.

\end{thebibliography}

%% ============================================================
%% APPENDIX
%% ============================================================
\appendix

\section{VLM Prototype Evaluation}\label{app:vlm}

We evaluated Qwen~3.5-397B-A17B, hosted with vLLM~\cite{kwon2023efficient} on 4$\times$H200 GPUs at FP8 quantization, and Gemini~3.1 Pro. Each model received one page image and the same structured prompt requesting blocks, signals, connections, and equations as JSON. Both models processed the same seeded sample of 100 pages; Qwen additionally processed all 6{,}979 pages.

Extracted entities were checked against the CSV-derived graph. We report precision only because the export records cannot be aligned completely to individual PDF pages; recall is therefore unknown. Connection precision uses relaxed matching on (signal, block, direction) without requiring an exact pin name. Equation accuracy requires an exact equation match after scoping the block by program and task metadata from the page footer. This evaluation cannot establish total extraction quality, but it is sufficient to quantify false positives in the rejected prototype.

Observed failures included fully qualified visual tags substituted for internal signal names, constants and pin labels emitted as signals, omitted negation operators, dropped equation terms, incorrect program/task metadata, and connections lost at page boundaries. Table~\ref{tab:vlm-comparison} reports the complete results.

\begin{table*}[t]
  \centering
  \caption{Direct VLM extraction from engineering-document pages. Sample columns use the same 100 seeded pages. Qwen Full covers all 6{,}979 pages; Gemini full-document time, token, and cost values are extrapolated from the sample ($\dagger$), while accuracy is not extrapolated. Recall is unavailable because export records cannot be aligned completely to individual pages.}
  \label{tab:vlm-comparison}
  % \small
  \begin{tabular}{lrrrr}
    \toprule
    & \multicolumn{2}{c}{Qwen~3.5-397B-A17B} & \multicolumn{2}{c}{Gemini~3.1 Pro} \\
    \cmidrule(lr){2-3} \cmidrule(lr){4-5}
    Metric & Sample & Full & Sample & Full$^\dagger$ \\
    \midrule
    Pages processed        & 100   & 6{,}979 & 100   & 6{,}979 \\
    Block precision        & 0.95  & 0.97    & 0.90  & --- \\
    Signal precision       & 0.31  & 0.33    & 0.34  & --- \\
    Connection precision   & 0.41  & 0.51    & 0.24  & --- \\
    Equation accuracy      & 75\%  & 36\%    & 4\%   & --- \\
    \midrule
    Wall time              & 3.0 min & 3.1 h  & 4.0 min & 4.6 h$^\dagger$ \\
    Input tokens / page    & \multicolumn{2}{c}{8.8K} & \multicolumn{2}{c}{1.7K} \\
    Output tokens / page   & \multicolumn{2}{c}{0.9K} & \multicolumn{2}{c}{2.5K} \\
    Total input tokens     & 0.9M  & 61M     & 0.2M  & 12M$^\dagger$ \\
    Total output tokens    & 0.1M  & 6M      & 0.3M  & 18M$^\dagger$ \\
    Est.\ cost             & ---   & \textasciitilde\$124\textsuperscript{a} & --- & \textasciitilde\$240\textsuperscript{b} \\
    \bottomrule
  \end{tabular}

  \smallskip
  \footnotesize\textsuperscript{a}4$\times$H200 GPUs at \$10/GPU-hour for 3.1 hours.
  \textsuperscript{b}Gemini~3.1 Pro at \$2/M input and \$12/M output tokens.
  $^\dagger$Extrapolated from the 100-page sample.
\end{table*}

\section{Query Evaluation Protocol}\label{app:query-eval}

We generated 40 questions using eight templates with five samples each. The templates cover signal producers and consumers, block inputs and outputs, upstream and downstream traces, and paths between named nodes. Samples were drawn from a bidirectional BFS neighborhood of the \texttt{L86CBPS\_F\_A} alarm containing 6{,}860 nodes and 14{,}056 edges. Query-type pruning produced an expected node-and-edge subgraph for each question; across the benchmark, retrieval operations returned 5--50 nodes and 5--100 edges.

Text references were generated from each question-specific expected subgraph using the DeepEval \texttt{Synthesizer}. The synthesis criteria required each reference to include all signals, blocks, directions, relationships, and available equations needed to answer the question while excluding unrelated structure. DeepEval \texttt{GEval} then compared each answer with both the reference text and graph context. Its rubric checked for contradictions, missing entities or relationships, incorrect path direction, false claims that no path exists, and irrelevant details; scores were normalized to 0--100\%.

Both pipelines used the same system instruction to answer only from retrieved control logic evidence. The agent could call block lookup, block input/output, signal producer/consumer, bounded path-tracing, and arbitrary-Cypher tools. It autonomously selected tools and stopping points. The fixed pipeline used the classification, resolution, and execution stages described in Section~\ref{sec:query}. This protocol uses generated questions, generated reference prose, and an LLM judge; the resulting correctness score should therefore be interpreted as a comparative development measure rather than an independently validated estimate of field accuracy.

\section{Worked Use-Case Example}\label{app:usecase}

This appendix presents the control logic query summarized in Section~\ref{sec:incident-replay}: the user's natural-language question and the system's response. Figure~\ref{fig:l86cbps} shows the returned dependency trace.

\smallskip\noindent\textbf{\underline{User query}:}
{\small\ttfamily What signals and conditions produce \textbf{L86CBPS\_F\_A}?}

\smallskip\noindent\textbf{\underline{Response}:}
{\small\ttfamily\noindent
The signal \textbf{L86CBPS\_F\_A} on device \textbf{G1} is produced through the following logic chain:
\begin{enumerate}[nosep, leftmargin=*]
  \item \textbf{Signal Flow:} \textbf{L86CBPS\_F\_A} is generated by block \textbf{MOVE\_13} from signal \textbf{L86CBPS\_F}; \textbf{L86CBPS\_F} by \textbf{MOVE\_3} from \textbf{L86CB1PS\_F}; and \textbf{L86CB1PS\_F} by \textbf{RUNG\_29} using equation \texttt{(A+D)*\textasciitilde B*C}.
  \item \textbf{Activation Conditions:} for \textbf{L86CBPS\_F\_A} to be true, \texttt{(A+D)*\textasciitilde B*C} must evaluate true --- i.e., the RUNG\_29 inputs must satisfy:
\end{enumerate}
\begin{itemize}[nosep, leftmargin=3em]
  \item at least one of A = 90CB1.POS\_CV\_DH or D = L86CB1PS\_F is \textbf{true};
  \item B = L14HS\_CBV (from L14HS) must be \textbf{false}, so \texttt{\textasciitilde}B is true; and
  \item C = L14HT\_CBV (from L14HT) must be \textbf{true}.
\end{itemize}
\noindent\textbf{Upstream Terminal Signals:} 34 signals including tank level and pressure indicators (TNK14HS2\_SEL, TNK14HR1, TNK14HP1, LK14PR1, TNK14LR1, etc.).
In summary, \textbf{L86CBPS\_F\_A} triggers when the high-temperature condition L14HT is active (C), the high-speed condition L14HS is inactive (\texttt{\textasciitilde}B), and either the valve position-deviation input or its latch is active (A+D).%
\par\noindent\textbf{Control Logic:} \underline{Trace \#1: G1$\mid$L86CBPS\_F\_A}}
\smallskip

Users can inspect the control logic trace interactively in the deployed interface. Figure~\ref{fig:l86cbps} shows a simplified version of the returned subgraph; the interface provides additional metadata and supports follow-up queries.

\end{document}